\documentclass{article}

\usepackage[nonatbib, preprint]{neurips_2022}

\usepackage[utf8]{inputenc} 
\usepackage[T1]{fontenc}    
\usepackage{hyperref}       
\usepackage{url}            
\usepackage{booktabs}       
\usepackage{amsfonts}       
\usepackage{nicefrac}       
\usepackage{microtype}      
\usepackage{xcolor}         

\usepackage{bbm, amsmath}
\usepackage{graphicx, wrapfig}
\usepackage{caption}
\usepackage{subcaption}
\usepackage{multirow}
\usepackage[flushleft]{threeparttable}

\newcommand\tstrut{\rule{0pt}{2.4ex}}
\newcommand\bstrut{\rule[-1.0ex]{0pt}{0pt}}

\title{DeepJoint: Robust Survival Modelling\\Under Clinical Presence Shift}

\author{%
Vincent Jeanselme\\
MRC Biostatistics Unit\\
University of Cambridge\\
\texttt{vincent.jeanselme@mrc-bsu.cam.ac.uk} \\
\And
Glen Martin\\
Health e-Research Centre\\
University of Manchester\\
\texttt{glen.martin@manchester.ac.uk} \\
\AND
Niels Peek \\
Health e-Research Centre\\
University of Manchester\\
\texttt{niels.peek@manchester.ac.uk} \\
\And
Matthew Sperrin \\
Health e-Research Centre\\
University of Manchester\\
\texttt{matthew.sperrin@manchester.ac.uk} \\
\AND
Brian Tom \\
MRC Biostatistics Unit\\
University of Cambridge\\
\texttt{brian.tom@mrc-bsu.cam.ac.uk} \\
\And
Jessica Barrett \\
MRC Biostatistics Unit\\
University of Cambridge\\
\texttt{jessica.barrett@mrc-bsu.cam.ac.uk}
}

\begin{document}

\maketitle

\begin{abstract}
Observational data in medicine arise as a result of the complex interaction between patients and the healthcare system. The sampling process is often highly irregular and itself constitutes an informative process. When using such data to develop prediction models, this phenomenon is often ignored, leading to sub-optimal performance and generalisability of models when practices evolve. We propose a multi-task recurrent neural network which models three clinical presence dimensions -- namely the longitudinal, the inter-observation and the missingness processes -- in parallel to the survival outcome. On a prediction task using MIMIC III laboratory tests \cite{johnson2016mimic}, explicit modelling of these three processes showed improved performance in comparison to state-of-the-art predictive models (C-index at 1 day horizon: 0.878). More importantly, the proposed approach was more robust to change in the clinical presence setting, demonstrated by performance comparison between patients admitted on weekdays and weekends. This analysis demonstrates the importance of studying and leveraging clinical presence to improve performance and create more transportable clinical models. 
\end{abstract}

\section{Introduction}
\label{sec:intro}
Machine learning has increasingly been leveraged in the medical context to model diagnoses and health outcomes given different data modalities \cite{katzman2018deepsurv, lee2018deephit, pmlr-v146-nagpal21a, jeanselme2022neural}. Despite promising results, medical data present additional domain-specific challenges that impact upon the applicability of machine learning in real-world settings. The complex interaction between patients and the healthcare system results in a dual representation of one's health status: times and types of observations not only reflect the patient's condition but the clinical staff's assessment of what was needed and when. Therefore, the sampling process is an informative process, which we refer to as \textbf{clinical presence} for which ordering, missingness, and observation times are informative proxies.

While the machine learning community has explored individual aspects of clinical presence, there are few clinical predictive models which explicitly model the full observation process. This limited research of clinical presence leads to approaches that assume non-informativeness. Although practitioners justify ignoring clinical presence as a way to avoid leveraging a changeable and unpredictable process \cite{van2020cautionary}, no evidence has been provided to justify this~\cite{lipton2016directly}.

Finally, a key challenge of clinical presence is its heterogeneity: the same underlying covariate distribution might present itself in different ways under different observation processes. For instance, note how change may occur between countries and over time due to new medical discoveries or insurance policies. The machine learning literature has explored how models may suffer from transportability under different labels, covariates, or concept shift \cite{zhang2013domain, moreno2012unifying}.  In this paper, we explore how a shift in the observation process might impact performance and how explicit modelling of clinical presence might offer more robust medical models.

\section{Related work}
\subsection{Clinical presence}
Medical observations reflect the complex interaction between patient, clinical staff, and healthcare services. In the context of a hospital stay, each observation from admission to discharge emerges from one of these interactions. If, when, and where patients seek care is not only linked to their medical condition but also their socio-economic characteristics, which will influence the care they can afford \cite{smith2018access}, the distance they will have to travel, and times they can access these services \cite{barik2015issues}. On the clinical side, the choice and timing of testing and treatment procedures are guided by the staff's assessment of the patient's condition. Clinical presence, therefore, breaks a key assumption of traditional statistical modelling: observations under clinical presence are an informative sample of the underlying distribution. Formally, for an unknown distribution $f(X)$, clinical presence consists in an observation process $o: X \rightarrow X_{obs}$ resulting in an observed distribution $f(X_{obs})$. Modelling $f(X_{obs})$ is an obstacle to the application of machine learning in healthcare \cite{xiao2018opportunities} as, under clinical presence, the observed distribution might not present the characteristics of the true distribution we aim to model, $f(X_{obs}) \neq f(X)$.

In this work, we look at how clinical presence might affect Electronic Health Records (EHRs) in three different ways and how to model them appropriately. First, observation times potentially provide information on the clinical assessment of a patient's condition. For instance, testing frequency may decrease when clinical staff believe that the patient's condition is improving. Second, the order of tests reflects their interpretation by clinical staff, e.g.~if a laboratory test is inconclusive, a second would follow. Third, missingness accompanies this previous process, e.g.~a missing value reflects its non-importance for diagnosis, or its non-informative value \cite{wells2013strategies}. These different dimensions have independently shown predictive edge \cite{lipton2016directly, groenwold2020informative, saar2007handling} and reduced biases emerging from this process~\cite{sperrin2017informative}. For instance,~\cite{agniel2018biases} shows improved 3-year-survival predictive performance when inputs contain missingness indicators, and \cite{lipton2016directly} discusses how using this information may improve mortality prediction. Our work shows how modelling them jointly may improve predictive performance and robustness to transfer by encoding clinical presence in the shared representation.

\subsection{Handling clinical presence}
Clinical presence has been handled in multiple ways in the statistical literature \cite{sisk2020informative}. In this work, we divide the methodologies used in the machine learning literature into four groups:
\begin{itemize}
\item Assumed uninformative
\item Leveraged for pre-processing
\item Featured
\item Jointly modelled
\end{itemize}

\paragraph{Assumed uninformative} One may assume that the observed distribution $f(X_{obs})$ is representative of the underlying distribution $f(X)$. For example, regular re-sampling or imputation assuming missing-at-random patterns rely on this hypothesis. While one could obtain equal fit with these approaches~\cite{molenberghs2008every}, inappropriate assumptions might lead to biased estimates and potentially lead to misleading conclusions. 

\paragraph{Leveraged for pre-processing} One can consider clinical presence informative of the longitudinal process and use pre-processing methods which model the clinical process to recover $f(X)$ from the observed distribution. Therefore, these approaches make assumptions on the observation process. However,~\cite{lipton2016directly} underlines how difficult it might be to adequately model clinical presence and questions the utility of these methodologies as they only partially recover the original distribution.

\paragraph{Featurized} One may consider clinical presence informative of the outcome of interest and therefore extracts measures of clinical presence reflecting medical practitioners' expertise that are then leveraged as model inputs. For instance, adding missing indicators \cite{lipton2016directly}, observation times \cite{agniel2018biases, che2018recurrent, sousa2020improving}, inter-observation time \cite{choi2016doctor, moskovitch2015outcomes}, or frequency \cite{pivovarov2014identifying} have improved models' performance. Similarly, in an automatic fashion, attention-based models leverage the observation times as inputs \cite{cai2018medical, choi2016retain, zhang2019attain}.

\paragraph{Jointly modelled} Instead of the two-step approaches described above, one can jointly model the clinical presence. In the statistical literature, joint models \cite{sperrin2017informative, gasparini2020mixed, su2019sensitivity}, marked point process \cite{islam2017marked} and Markov models \cite{alaa2017learning} have been proposed to model the visit process and the outcome of interest. However, these methods either rely on strong parametric assumptions or do not scale to large datasets. Machine learning models have also been extended to leverage missing data \cite{twala2008good} and recurrent models to handle irregularities in time series \cite{weerakody2021review}. The key idea of this last family of methods is to integrate the temporal information by slowly converging to a `stable' latent state as time passes through the use of a decay parameter \cite{che2018recurrent, baytas2017patient, pham2016deepcare} or multi-level memory \cite{mozer2017discrete}. For instance, GRU-D \cite{che2018recurrent} has shown promising results by extending Gated Recurrent Unit (GRU) with an exponential decay on the hidden state, incorporating the temporal dimension in the embedding. 

While all these approaches aim to tackle the problem of clinical presence, they make different assumptions on its informativeness, have usually focused on one specific aspect of it, and didn't study the impact of change in this observation process.

\subsection{Robustness to clinical presence shift}
A fundamental feature of clinical presence is its propensity to vary as medical practices might differ across hospitals, regions and countries and be subject to policies, insurance incentives and the evolution of medical knowledge. This shows a crucial limitation of the application of machine learning for medical data as models may not generalise and may become outdated as practices and policies evolve \cite{van2020cautionary}. Evidence of this problem has been suggested by region and time performance shift \cite{guo2022evaluation, lazer2014parable, nestor2019feature, singh2022generalizability}.

Shifts in time or between hospitals have been characterised and studied as covariates, labels or concept shifts \cite{zhang2013domain, moreno2012unifying}. Given a set of covariates $X$ and an outcome $Y$, a covariates (resp. labels) shift consists of a shift in $f(X)$ (resp. $f(Y)$), and concept shift results in a change in the relationship between input and output, i.e. $\mathbbm{P}(Y|X)$. Detecting and measuring shifts have led to methodologies to alleviate these phenomena \cite{lu2018learning, lipton2018detecting}, and to more robust clinical embeddings \cite{guo2022evaluation, nestor2019feature, rajkomar2018scalable}.

Clinical presence shift can be understood as a combination of both covariate and concept shifts with regard to the observed distribution $f(X_{obs})$. Assuming the underlying distribution stable, a shift in the observation process $o$ to $o'$ results in a shift in the available data $X_{obs}$, i.e. a covariate shift from $f(o(X))$ to $f(o'(X))$. Additionally, despite the preservation of $\mathbbm{P}(Y|X)$, the modelled $\mathbbm{P}(Y|X_{obs})$ also suffers from a shift. We propose to explicitly model clinical presence in a multi-task setting to detect this potential shift in the observation process and regularise the model to obtain a more robust data embedding. We show how this approach results in improved performance and more robustness to shift in clinical presence.

\section{Deep Joint}
In this paper, we study the problem of clinical presence in the context of EHRs. More specifically, we focus on laboratory tests after admission to the Intensive Care Unit (ICU) to predict patient survival as these tests are ordered by clinical staff at specific times and will be more informative in their sampling patterns than semi-automatic measurements. We propose a recurrent neural network that jointly models the observation process, namely: (i) the time of the next observation using a temporal point process neural network, (ii) the presence of specific tests and (iii) the expected results using neural networks with Bernoulli and Gaussian likelihoods respectively. Finally, an extension of Cox regression leveraging a neural network, known as DeepSurv \cite{katzman2018deepsurv}, models the survival outcome. The proposed model allows scalable joint modelling by an end-to-end gradient descent maximisation of the full likelihood.

\subsection{Notations}
The studied medical setting consists of a population of $N$ individuals with a vector of laboratory tests $lab_{i,j}$ -- potentially with missing values -- observed through time, in which the notation $j \in [\![0, l_i]\!]$ denotes the $j^{th}$ observation for patient $i$ at time $t_{i,j} \in \mathbbm{R}^+$, where $t_{i, 0} = 0$ is the patient's admission to the ICU and $l_i$ is the last observation for patient $i$. The proposed approach focuses on modelling the clinical presence aspect of these data. To do so, we introduce the two following notations. First, we model the elapsed time between two observations: $\epsilon_{i,j} = t_{i,j} - t_{i, j-1}$, note the subscript $i$ underlines that observations may not happen at similar times for different patients. Second, the observation process is also characterised by which laboratory tests are performed. Consequently, we introduce $m_{i, j} = [\mathbbm{1}_{\text{Lab test } k \text{ observed at } j}]_{k \in \text{labs}}$  a vector of indicators of observed laboratory tests at observation time $t_{i,j}$.

Finally, the aim is to model the survival outcome after the first 24 hour of observation. Each patient has, therefore, an associated time of the end of follow-up $T_i$ and the type of observed event $d_i$ for which $d_i = 1$ signifies that the patient died, $d_i = 0$ means the patient was censored.

\subsection{Architecture}
\label{sec:methodology}
\begin{figure}[ht]
\centering
\includegraphics[width=\linewidth]{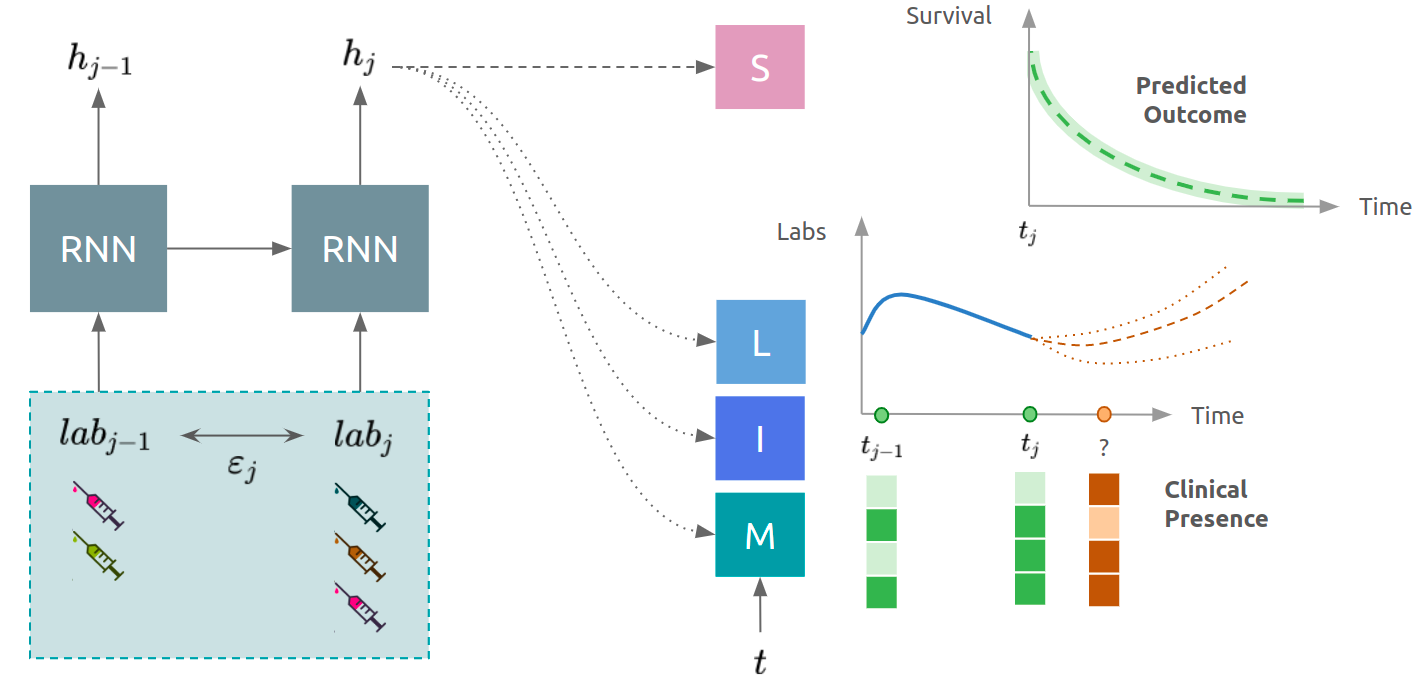}
\label{fig:model}
\caption{Deep Joint Model - Multi-task modelling of clinical presence and survival outcome.\newline\textit{An LSTM network extracts an embedding \textit{h} which is leveraged to model the clinical presence mechanisms through the networks \textbf{L} -- to model the longitudinal evolution, \textbf{I} -- to model the inter-observation time, \textbf{M} -- to model the missingness patterns, and the survival outcome through the network \textbf{S}}.}
\end{figure} 

The proposed architecture decomposes clinical presence into three aspects (see Section \ref{sec:intro}) that are modelled through three different neural networks relying on the embedding $h_{i, j}$ outputted by a Recurrent Neural Network (RNN) at the observation time $t_{i,j}$. To extract this embedding, we adopted a Long Short Term Memory (LSTM) network \cite{hochreiter1997long} that leverages the sequential nature of the observed data and can capture temporal dependencies. Note, however, that the inputs are the series of observations at irregular times. The resulting embedding is then leveraged to model the survival outcome. We now introduce the networks used for each of these tasks.

\paragraph{Longitudinal process} Similarly to traditional joint models, the latent state $h_{i,j}$ is leveraged through a multi-layer perceptron $L$ to model the future value of each laboratory test to be observed after a period $t$ from the prediction time $t_{i,j}$. Assuming a Gaussian distribution for the observed data, the neural network $L$ models the mean and variance of each future covariate given both the embedding and the prediction time, i.e. 
$$\widehat{lab}_i(t) \sim \mathcal{N}(\mu_{i}(t), \sigma_{i}^2(t))$$

with $\mu_{i}(t)$ and $\sigma_{i}(t)$ the means, variances vectors outputted by $L(h_{i, j}, t)$.

\paragraph{Missingness process} $h_{i,j}$ is also used in a second multi-layer perceptron $M$ to model the likelihood of observing the different covariates, i.e. which tests are likely to be performed after the period $t$. One can assume a Bernoulli distribution over the missingness patterns, i.e.
$$\hat{m}_i(t) \sim \mathcal{B}ern(M(h_{i,j}, t))$$

\paragraph{Temporal process} A monotonic positive neural network $I$ leverages $h_{i,j}$ to model the recurrent patterns of observations. This neural network with constrained positive weights models the likelihood of observing any new laboratory test after time $t$ following the approach proposed in \cite{xiao2017modeling}. The model allows the computing of the risk of observing an event at any time  $\Delta$ after $t$ minutes by directly outputting the cumulative hazard $I$ while avoiding any parametric assumption, i.e.
$$\mathbbm{P}(\Delta > t) = \exp(- I(h_{i,j}, t))$$

\paragraph{Survival outcome} Finally, to model the hazard function $\lambda(t)$ defining the survival model, we used the DeepSurv model \cite{katzman2018deepsurv} that leverages a multi-layer perceptron $S$ to extract a non-linear covariates shift used in a standard multiplicative proportional hazards Cox model. Given $\lambda_0$, the baseline hazard, the hazard is expressed as:
$$\lambda(t) = \lambda_0(t) \exp(S(h_{i,j})) $$

The final model, therefore, models both clinical presence and survival outcome by leveraging the representation $h_{i,j}$ extracted by the LSTM. This embeds the observation process in the latent representation leveraged for survival modelling.

\subsection{Training}
For training, the aim is to maximise the likelihood of observations while maximising the survival likelihood. To do so, the model is trained in a multi-task fashion in which each objective is back-propagated simultaneously by averaging the loss of the different tasks: survival, longitudinal, timing and missingness processes. As balancing the losses may greatly impact performance \cite{gong2019comparison}, we ensured that no clinical presence task was over-represented in the average by computing negative log-likelihoods for each task and using a dynamic weighting average scheme \cite{liu2019end}. This consists in weighting each loss related to clinical presence by its relative change at iteration $e$ as follows:
$$\forall\;task \in \{L, I, M\}, w_{task}(e) = \log \frac{L_{task}(e)}{L_{task}(e - 1) * \theta}$$ which is then normalised between the three tasks using a Softmax. $L$ is the average training likelihood for the given task and $\theta$ is a temperature hyperparameter that controls softness, i.e. larger values would lead to equal weights.
\paragraph{Longitudinal process} This first neural network aims to maximise the likelihood of the next \textit{observed} laboratory tests values at time $\epsilon_{i, j}$. The assumption of normality leads to the use of the Gaussian log likelihood loss:
\begin{align*}
    l_L =& \sum_i \sum_{j \in [\![1, l_i-1]\!]} m_{i, j+1} * \log \mathcal{N}(lab_{i, j+1} |\mu_{i, j}, \sigma_{i, j}^2)\\
    =& - \sum_i \sum_{j \in [\![1, l_i-1]\!]} m_{i, j+1} * \left(\frac{(lab_{i, j+1} - \mu_{i, j})^2}{2{\sigma_{i, j}}^2} + \log\sqrt{2\pi {\sigma_{i, j}}^2} \right)
\end{align*}
with $\mu_{i, j}$ and $\sigma_{i, j}$ the means, variances vectors outputted by $L(h_{i, j}, \epsilon_{i, j+1})$ and $*$ the element-wise multiplication. Note the filtering through $m_{i, j+1}$ to avoid a penalty on unobserved data.

\paragraph{Missingness process} Similarly, the loss for the missingness process relies on the Bernoulli distribution. This leads to the binary cross entropy loss:
\begin{align*}
    l_M =& \sum_i \sum_{j \in [\![1, l_i-1]\!]} \log \mathcal{B}ern(m_{i, j+1} | M(h_{i, j}, \epsilon_{i, j+1}))\\
    =& - \sum_i \sum_{j \in [\![1, l_i-1]\!]}  \left( m_{i, j+1} * \log[M(h_{i, j}, \epsilon_{i, j+1})] + (1 - m_{i, j+1}) * \log[1 - M(h_{i, j}, \epsilon_{i, j+1})] \right)
\end{align*}

\paragraph{Temporal process} The likelihood of the timing process can be exactly modelled without the need for distributional assumptions on the inter-observation distribution as the neural network returns the cumulative intensity at horizon $t$. Additionally, the instantaneous intensity can be computed by differentiation, at no extra computational cost in the context of neural networks training procedure.
$$l_I = \sum_i \sum_{j \in [\![1, l_i-1]\!]} \left( I(h_{i, j}, \epsilon_{i, j+1}) - \log \frac{dI(h_{i, j}, t)}{dt}_{t = \epsilon_{i, j+1}} \right)$$

\paragraph{Survival outcome} Finally, the survival model, DeepSurv, relies on the proportional hazards assumption made by the Cox model. This results in the following log likelihood:
$$l_S = \sum_{i, d_i = 1} \left( S(h_{i, j}) - \log \sum_{k, T_k > T_i} \exp(S(h_{i, j})) \right)$$
Then, the population baseline hazard, $\lambda_0(t)$, is estimated as piece-wise constant based on the training population.

The final loss is therefore defined at iteration $e$ as $$l(e)  = (1 - \alpha) l_S + \alpha \sum_{task \in\{L, I, M\}} w_{task}(e) * l_{task}(e)$$
With $\alpha$, a hyperparameter balancing between survival and clinical presence. This loss was evaluated on a validation set and an early stopping strategy was adopted to stop the multi-task training. Then, a fine-tuning of the network $S$ with all other weights fixed -- using the same stopping criterion -- allowed to improve the survival performance.


\section{Experiments}
\label{sec:experiments}
This paper explored the problem of in-hospital survival prediction after 24 hours of ICU admission using laboratory tests. We used data from the Medical Information Mart for Intensive Care III (MIMIC III) \cite{johnson2016mimic}. While the clinical actionability of these predictive models has been questioned \cite{schoenfeld2005survival}, it serves the optimisation of resource allocation and remains a standard comparison task in the machine learning literature. The code for the proposed models and experiments is available on Github\footnote{\url{https://github.com/Jeanselme/ClinicalPresence/}}.

\subsection{Data}
The MIMIC III dataset gathers anonymized laboratory tests, vital signs and diagnoses of 38,597 patients admitted to the Beth Israel Deaconess medical centre between 2001 and 2012. This analysis focused on the laboratory tests as their ordering will follow from medical observations and practitioners' expertise. While other measurements such as vital signs monitoring may improve outcome modelling, we chose to demonstrate the impact of clinical presence on a limited set of potentially informative covariates as other modalities might present different clinical presence patterns. After pre-processing \cite{wang2020mimic}, we selected -- using an ECLAT algorithm \cite{zaki1997parallel} -- a set of adults with shared laboratory tests for whom at least one laboratory test of each was performed during the 24 hours after admission and who survived at least this amount of time. This leads to a subset of 30,834 patients with 17 different laboratory tests (See Table \ref{tab:pop} in Appendix for a population description and Table \ref{tab:labs} for the list of laboratory tests). Note that similar results are observed when all laboratory tests are considered, but are more sensitive to the imputation strategies which are out of the scope of this analysis.

\subsection{Modelling clinical presence}
The proposed approach models the in-hospital survival of patients admitted to the hospital after 24 hours of observation. To measure the gain of leveraging clinical presence, we compared our method \textbf{DeepJoint} against multiple strategies relying on last observation carried forward data with patient-mean imputation, i.e. any missing laboratory tests is replaced by the last observed results and remaining missing data are imputed by the patient's mean per lab test type -- guaranteed to exist because of the selection of patients presenting each laboratory test at least once. 
First, we compared against two baseline approaches:
\begin{itemize}
    \item \textbf{Last observation (Last)}: Extract the last observation $l_i$ as representation for each patient. This approach assumes no informativeness in the evolution of the laboratory results.
    \item \textbf{Summarising (Count)}: Add to the previous representation the number of each test performed in the first 24 hours. This approach assumes informativeness of the counting process but ignores the laboratory test evolution.
\end{itemize}

Second, we used RNN-based approaches to take advantage of the longitudinal evolution of the laboratory tests:
\begin{itemize}
    \item \textbf{Ignoring clinical presence (Ignore)}: An LSTM is trained on the imputed data leveraging the inputs' temporal order but ignoring their irregularity and missingness patterns.
    \item \textbf{Resampling (Resample)}: Imputed data are re-sampled every hour to satisfy the LSTM regularity assumption. This assumes non informativeness of the sampling process for the outcome of interest.
    \item \textbf{Modelling (GRU-D)}: The imputed data concatenated with missingness indicators serve as inputs to a GRU-D model \cite{che2018recurrent} that models the inter-observation time as a decaying function, following the intuition that the hidden state should converge to a stable state with time.
    \item \textbf{Featurization (Feature)}: Imputed longitudinal data are concatenated with missingness indicators and time elapsed since previous observation \cite{lipton2016directly}, allowing the LSTM to extract information from these proxies of the clinical presence.
\end{itemize}

Finally, we propose two additional variants of our proposed model that aim to identify which aspect of clinical presence modelling provides a predictive and robustness edge:
\begin{itemize}
    \item \textbf{DeepJointFeature} which consists of using the same input as \textbf{Feature} to both model and leverage the information contained in clinical presence as input of the recurrent neural network.
    \item \textbf{DeepJointFineTune} for which the entire \textbf{DeepJointFeature} model is fine tuned to maximise the survival likelihood in the last 500 iterations. This approach only leverages clinical presence to initialise weights.
\end{itemize}

For all experiments, input data were normalised and we used DeepSurv to model the hazard shift from the population hazard baseline using a multi-layer perceptron (see Section \ref{sec:methodology}). All methods were trained using an Adam optimiser over 1000 epochs with early stopping on 10\% of the training set. 500 epochs were used to train the LSTM and any additional outcomes and the last iterations were used to fine tune the survival network only. Additionally, hyperparameter tuning was performed on a 10\% split of the remaining training set on the grid presented in Appendix Table \ref{tab:grid}.

Models were evaluated on a 90\%-10\% train-test patients split. Survival was inferred on the embedding obtained at the last observation $l_i$ in the 24 hour post-admission and performances were compared using time-dependent C-index \cite{hung2010estimation} and Brier score \cite{graf1999assessment} evaluated at time horizons of 1, 7 and 14 days after the observation period. 95\% confidence intervals were obtained using 100 bootstrapped iterations on the test set predicted values.

\subsection{Robustness to change in clinical presence}

\begin{figure}[ht]
\centering
    \begin{minipage}[c][100px][t]{.55\textwidth}
      \centering
      \includegraphics[height=65px]{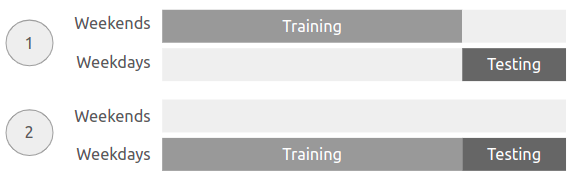}
      \captionof{figure}{Robustness evaluation between patients admitted on weekdays and weekends.}
      \label{fig:robust}
    \end{minipage}%
    \hspace{0.5cm}
    \begin{minipage}[c][100px][t]{.40\textwidth}
        \centering
      \includegraphics[height=75px]{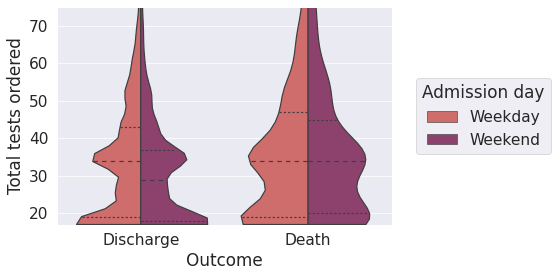}
      \captionof{figure}{Total number of tests performed in the 24 hours after admission to the ICU.}
      \label{fig:diff_weekday}
    \end{minipage}
\end{figure}

The previous comparison shows how handling clinical presence in different ways may impact predictive performance. However, as mentioned earlier, one characteristic of clinical presence is its heterogeneity: experts' practice might change from one country to another, from one hospital to another, or even in one hospital as practices change across specialities or exogeneous factors such as day of the week. In this second set of experiments, we focused on this last scenario. Motivated by the difference in the distribution of the numbers of tests performed during weekends and weekdays (see Figure \ref{fig:diff_weekday}), we hypothesised that practice might differ during weekends and weekdays admissions due to physicians and laboratory availability. 

The weekend effect is a well-known -- while not well-understood -- phenomenon in public health: patients admitted on weekends are more likely to die than patients on weekdays \cite{pauls2017weekend}. While providing an opportunity to study the transfer of survival models, this outcome difference represents a difference in label distributions for which performance cannot directly be compared. To make this comparison possible, we adopted the evaluation methodology used in the robustness literature. Patients were separated given days of admission (from Monday 8 am to Saturday 8 am vs the rest). Each group of admission were then further divided between training and testing. As described in Figure \ref{fig:robust}, a first model uses the train set of patients admitted on weekends and tested on the test set of weekdays-admitted patients. Then, a second model uses the train set of patients admitted on weekdays and tested on the test set of weekdays-admitted patients. The two resulting models are comparable as they are applied to the same test set. Moreover, to ensure that performance differences were not due to training sample size, the weekends' population was over-sampled to match the weekdays' one (similar results were observed when leveraging the whole available populations).

Finally, the opposite analysis was performed for the weekend test set. We compared the three best-performing methodologies and the \textbf{Last} baseline in this second set of experiments using identical performance metrics.

\section{Results}
\label{sec:results}
\subsection{Modelling clinical presence improves performance}
\begin{figure}[ht]
\centering
\includegraphics[width=\textwidth]{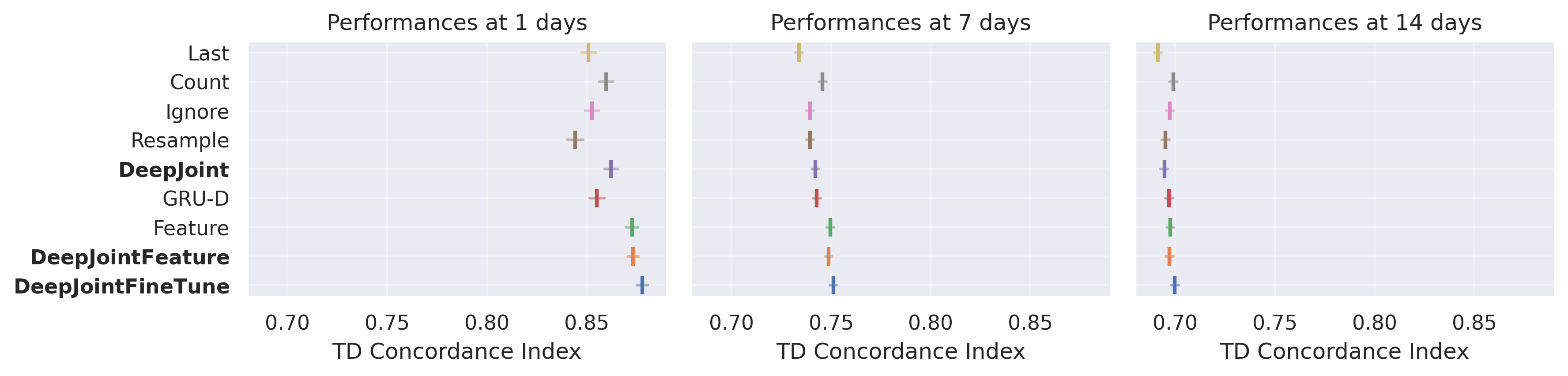}
\caption{C-Index performance on a random split of the population with 95\% confidence intervals\newline\textit{Higher is better -- The proposed methodologies (in bold) present state-of-the-art performance on short time horizons compared to models based on the same input. This advantage fades as clinical presence is informative of short-term instability.}}
\label{fig:random}
\end{figure}
Figure \ref{fig:random} describes the discrimination of the models at different evaluation horizons on the test set (see also Appendix \ref{app:results} for Brier score and detailed results). The three proposed approaches present competitive performances compared to models based on the same input. First, \textbf{DeepJoint} only based on the laboratory tests has improved C-index in comparison to \textbf{Ignore} or even \textbf{GRU-D} which leverages missingness as inputs. This result shows how modelling clinical presence, even if not leveraged as inputs, leads to more predictive embedding.

Next, note how the baseline method \textbf{Last} gains from extracting features reflecting the observation process (\textbf{Count}). Similarly, \textbf{DeepJointFeature} improves over \textbf{DeepJoint} and presents equivalent results to \textbf{Feature} showing that the joint modelling does not provide additional predictive improvement when informative features are fed as inputs and the model leverages the temporality of the data. Finally, a full fine-tuning of the network after modelling clinical presence provides an additional edge, suggesting that clinical presence initialisation leads to a more stable minimum.

Note how all methodologies have decreasing performance with larger time horizons. This observation results from the complexity of modelling long term survival for critically unstable patients. Additionally, the reduced edge of modelling clinical presence stresses how the observation process reflects short term instability as increased testing might indicate critical conditions. 

These experiments show how modelling and featuring of clinical presence improves discriminative performance. Nonetheless, we hypothesised that clinical presence modelling leads to more robust embedding when a shift in the observation process is observed.

\subsection{Joint modelling improves robustness}
\label{sec:rob}
\begin{figure}[ht]
\centering
\includegraphics[width=\textwidth]{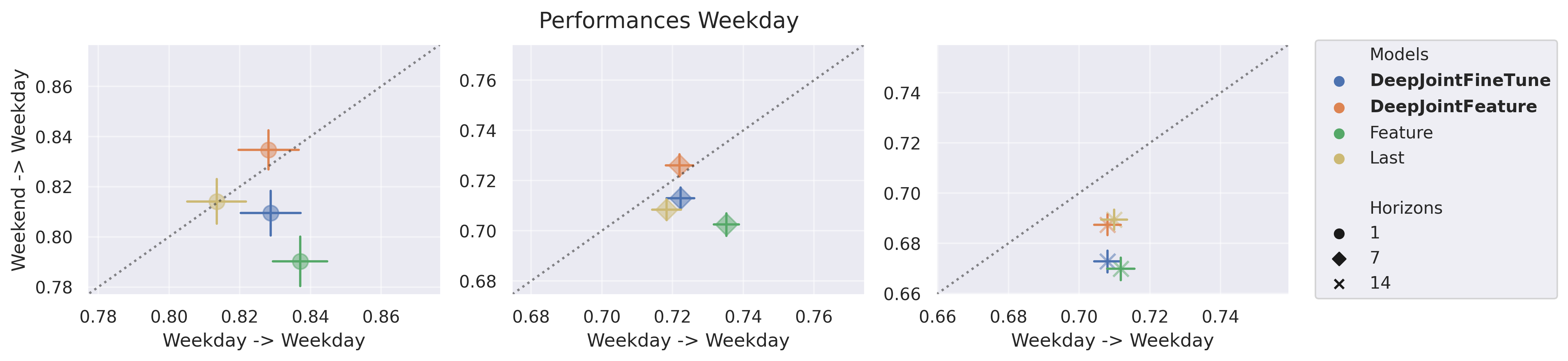}
\caption{Discriminative performance evaluated on patients admitted on weekdays for a transferred model (\textit{y-axis}) and an oracle model trained on the train weekdays-admitted patients and tested on the test set of this same group (\textit{x-axis}) (with 95\% confidence intervals). \newline\textit{Models closer to the diagonal are more robust to change in clinical presence -- \textbf{DeepJointFeature} presents better property under this clinical presence shift with robust discriminative performance.}}
\label{fig:split_weekday}
\end{figure}

Figure \ref{fig:split_weekday} presents the performance change of transferring the model from one clinical presence setting to another for the top performing proposed methodologies (see Appendix for detailed results Table \ref{tab:transfer} and the complementary scenario in Appendix Figure \ref{fig:split_weekend}). A model robust to change would be one for which the performance on the test set of the same setting as the training set is similar to the model that has been transferred, i.e. models close to the diagonal. Models under the diagonal show overfitting to the training set. Therefore, one should select a model both close to the diagonal, and with the strongest discriminative performance (upper right corner).

First, \textbf{Feature} presents the best predictive performance on the training set but is sensitive to the setting shift. At the opposite end, the model based on the last observation (\textbf{Last}) presents the worst discriminative performance. However, as the model does not leverage clinical presence, changes in the observation process impact its performance to a lesser extent. The proposed \textbf{DeepJointFeature} features similar level of robustness but with significantly improved discrimination. Conversely, despite presenting better predictive performances at the population level, the proposed fine-tuned variant is prone to overfitting. This underlines how modelling jointly the clinical presence regularises the embedding against the shift in the observation process. 

Similarly to the previous experiment, models' differences fade with larger time horizons. However, note how \textbf{DeepJointFeature} remains the model presenting the most robust performance. 

This experiment underlines how jointly modelling clinical presence and survival may improve the robustness of the embedding to change. However, traditional strategies of leveraging this information may appear to improve performance in internal validation but are prone to overfitting when transferred to a different clinical presence setting.

\section{Discussion}
\label{sec:discussion}
In this work, we propose to explicitly model the clinical presence reflected in the ordering, timing and missingness of observational longitudinal data. The novelty of the proposed approach does not only lie in its discriminative performance but in its transportability under change in the observation process compared to state-of-the-art approaches. Additionally, this modelling allows the detection of change in the observation process.

Approaching clinical presence as a multi-task problem can be understood as an embedding regularisation \cite{sagawa2019distributionally} which results in a more clinically relevant representation for survival prediction. However, despite multi-task improvements observed in the literature, theoretical foundations are still lacking \cite{ruder2017overview}. Another limitation of this work is the lack of consideration of potential competing risks, but also the assumption of independence between patients: in the medical context, and even more particularly in the ICU setting, prioritisation is key in the delivery of care. When considering clinical presence, the assumption of independence might consequently be questioned and calls for future work. While we propose a set of experiments to study robustness in the MIMIC III dataset, this focuses on one hospital, external validation would allow the quantification of the methodology's robustness under different shifts. Lastly, our modelling highlighted the importance of modelling clinical presence in the context of laboratory tests. Studying the relation between adding new covariates and the change in predictive performance would indicate potential correlation between covariates and medical behaviour.

Finally, we would like to echo the remark on the risk of leveraging clinical presence information raised by~\cite{van2020cautionary, lipton2016directly} in which the authors underline how taking advantage of missing data might lead to a mismatch between the training and deployment settings, but that ignoring this information might not be possible. Our work proposes a way to leverage this information while informing the user of changes in the patterns of observation. Nonetheless, practitioners should be wary of informing medical practice by leveraging clinical presence as it may result in a self-fulfilling prophecy and reinforced biases. Further work on the impact on medical behaviour would be a necessary step towards deployment.

\newpage
\begin{ack}
This work has been partially funded by UKRI Medical Research Council (MC\_UU\_00002/5 and MC\_UU\_00002/2).
\end{ack}

\bibliographystyle{ieeetr}
\bibliography{bibabstract}

\newpage
\section*{Checklist}

\begin{enumerate}

\item For all authors...
\begin{enumerate}
  \item Do the main claims made in the abstract and introduction accurately reflect the paper's contributions and scope?
    \answerYes{See Section~\ref{sec:results}.}
  \item Did you describe the limitations of your work?
    \answerYes{See Section~\ref{sec:discussion}.}
  \item Did you discuss any potential negative societal impacts of your work?
    \answerYes{See last paragraph of Section~\ref{sec:discussion} on the risk of reinforcing error.}
  \item Have you read the ethics review guidelines and ensured that your paper conforms to them?
    \answerYes{}
\end{enumerate}

\item If you ran experiments...
\begin{enumerate}
  \item Did you include the code, data, and instructions needed to reproduce the main experimental results (either in the supplemental material or as a URL)?
    \answerYes{Code will be released upon acceptance of the manuscript and have been shared as a supplemental material.}
  \item Did you specify all the training details (e.g., data splits, hyperparameters, how they were chosen)?
    \answerYes{See Section~\ref{sec:experiments} and Appendix \ref{app:experiments}.}
	\item Did you report error bars (e.g., with respect to the random seed after running experiments multiple times)?
    \answerYes{Bootstrapped confidence intervals are reported.}
	\item Did you include the total amount of compute and the type of resources used (e.g., type of GPUs, internal cluster, or cloud provider)?
    \answerYes{See Appendix \ref{app:experiments}.}
\end{enumerate}

\item If you are using existing assets (e.g., code, data, models) or curating/releasing new assets...
\begin{enumerate}
  \item If your work uses existing assets, did you cite the creators?
    \answerYes{All models and data source were cited in Section~\ref{sec:experiments}.}
  \item Did you mention the license of the assets?
    \answerNo{References are included to access license details.}
  \item Did you include any new assets either in the supplemental material or as a URL?
    \answerYes{All code is publicly available under a MIT licence.}
  \item Did you discuss whether and how consent was obtained from people whose data you're using/curating?
    \answerNo{All data and code used were publicly available and received prior IRB approval.}
  \item Did you discuss whether the data you are using/curating contains personally identifiable information or offensive content?
    \answerYes{All data were anonymized as mentioned in Section~\ref{sec:experiments}.}
\end{enumerate}

\end{enumerate}

\newpage
\appendix

\section{Appendix}

\subsection{Experiments}
\label{app:experiments}

\subsubsection{Data characteristics}
Table \ref{tab:pop} presents the demographic characteristics of the studied population and \ref{tab:labs} summarises the set of tests selected with the mean number of tests performed during the 24 hours post admission and their mean values.
\begin{table}[ht]
    \centering
    \begin{threeparttable}
        \caption{Population characteristics between patient admitted on weekdays and weekends.}
        \label{tab:pop}
        \begin{tabular}{ccrccc}
        &&& \multicolumn{3}{c}{Population}\bstrut\\\cline{4-6}
          &    &      &  Overall &  Weekday &  Weekend \tstrut\bstrut \\
            \hline\hline
            \multicolumn{3}{r}{Number of patients} &30,834& 23,359 & 7,475 \bstrut\tstrut\\
            \hline
            \multirow{5}{*}{Outcome} & \multicolumn{2}{r}{Length of stay (in days$^*$)} & 10.05 (10.49) &    10.05 (10.60) &    10.03 (10.15)\tstrut\bstrut\\\cline{3-6}
            &\multirow{4}{*}{Death} &Overall (\%)         &       13.86 &    13.50 &    15.00 \tstrut\\
            && 1 day$^+$ (\%) &       1.15 &    1.13 &    1.20  \\
            && 7 days$^+$ (\%) &       7.05 &    6.79 &    7.88 \\
            && 14 days$^+$ (\%)  &       10.33 &    9.94 &    11.52 \bstrut \\
            \hline
            \multirow{10}{*}{Demographics} 
            &\multirow{2}{*}{Gender} & Male (\%) &       56.43 &    56.38 &    56.62 \tstrut\\
             && Female (\%)&       43.57 &    43.62 &    43.38 \bstrut\\
            \cline{3-6}
            &\multirow{5}{*}{Ethnicity} & White (\%)&        71.84 &    72.52 &    69.74 \tstrut\\
             & & Other (\%) &       14.52 &    14.03 &    16.04 \\
              && Black (\%)&        7.88 &     7.87 &     7.92 \\
             & & Hispanic (\%)&        3.32 &     3.19 &     3.73 \\
             & & Asian (\%)&        2.43 &     2.39 &     2.57 \bstrut\\
            \cline{3-6}
           & \multirow{3}{*}{Insurance} & Public (\%)&       66.15 &    65.68 &    67.59 \tstrut\\
             & & Private (\%) &       32.65 &    33.31 &    30.58 \\
             & & Self Pay (\%)&        1.21 &     1.01 &     1.82 \\
        \end{tabular}
        \begin{tablenotes}
            \small
            \item $^*$ Mean (std)
            \item $^+$ After first day of observation
        \end{tablenotes}
      \end{threeparttable}
\end{table}

\begin{table}[ht]
    \centering
    \begin{threeparttable}
        \caption{List of laboratory tests used with associated mean number of tests and values (and standard deviations).}
        \label{tab:labs}
        \begin{tabular}{rcc}
            Laboratory test & Number of tests & Value \bstrut\\
            \hline\hline
            Anion gap                                 &        1.84 (1.00) &  13.87 (3.23) \tstrut\\
            Bicarbonate                               &        1.88 (1.00) &  24.15 (4.34) \\
            Blood urea nitrogen                       &        1.90 (1.00) &  24.91 (20.36) \\
            Chloride                                  &        1.92 (1.03) & 105.09 (5.74) \\
            Creatinine                                &        1.91 (1.00) &   1.33 (1.38) \\
            Glucose                                   &        2.84 (2.59) & 136.77 (49.89) \\
            Hematocrit                                &        2.89 (2.36) &  32.89 (5.34) \\
            Hemoglobin                                &        2.39 (2.04) &  11.09 (1.91) \\
            Magnesium                                 &        1.81 (0.97) &   1.98 (0.37) \\
            MCH$^*$                                   &        1.76 (0.96) &  30.29 (2.52) \\
            MCH concentration                         &        1.76 (0.96) &  33.82 (1.52) \\
            Mean corpuscular volume                   &        1.76 (0.96) &  89.63 (6.65) \\
            Platelets                                 &        1.88 (1.10) & 224.07 (109.73) \\
            Potassium                                 &        2.01 (1.08) &   4.13 (0.55) \\
            Red blood cell count                      &        1.87 (1.02) &   5.23 (15.24) \\
            Sodium                                    &        1.93 (1.08) & 138.89 (4.38) \\
            White blood cell count                    &        1.88 (1.03) &  11.83 (13.08) \\
        \end{tabular}
        \begin{tablenotes}
            \small
            \item $^*$ Mean corpuscular hemoglobin.
        \end{tablenotes}
      \end{threeparttable}
\end{table}

\newpage
\subsubsection{Hyperparameters tuning}
All models' hyper-parameters were selected over the following grid of hyperparameters (if appropriate) following 100 iterations of random search. All experiments were ran on a A100 GPU over 100 hours for each experiment.

\begin{table}[ht]
    \centering
    \begin{threeparttable}
        \caption{Grid for hyperparameters search}
        \label{tab:grid}
        \begin{tabular}{crl}
                                  & Hyperparameter        & Values                         \bstrut                \\
                                  \hline\hline
        \multirow{4}{*}{Training}                  & Learning rate         & $10^{-3}$, $10^{-4}$ \tstrut\\
                                  & Batch size            & $100$, $250$                                       \\
                                  & $\alpha$              & $0.1$, $0.5$                                       \\
                                  & $\theta$                     & $2^*$                                   \bstrut           \\
                                  \hline
        \multirow{2}{*}{RNN}                       & Layers                & $1$, $2$, $3$                      \tstrut                  \\
                                  & Hidden nodes          & $10$, $30$                                       \bstrut  \\
                                  \hline
        \multirow{2}{*}{Survival}                  & Layers                & $0$, $1$, $2$, $3$                   \tstrut                  \\
                                  & Nodes                 & $50$                                            \bstrut \\
                                  \hline
        \multirow{3}{*}{Clinical Presence}
                                  & Longitudinal          & Same parameters explored as survival         \tstrut  \\
                                  & Temporal              & Same parameters explored as survival           \\
                                  & Missing               & Same parameters explored as survival           \\
        \end{tabular}
        \begin{tablenotes}
            \small
            \item $^*$ Following the results from \cite{liu2019end}.
        \end{tablenotes}
      \end{threeparttable}
\end{table}

\newpage
\subsubsection{Results}
\label{app:results}

Table \ref{tab:res} presents the discriminative results on the test set of the first set of experiment for which patients are randomly assigned to train and test sets.
\begin{table}[ht!]
    \centering
    \caption{Time Dependent C-Index - Mean (95\%-CI) -- \textit{Higher is better.}}
    \label{tab:res}
    \begin{tabular}{rccc}
    {} & \multicolumn{3}{c}{Evaluation horizons}\\
    {} & \multicolumn{3}{c}{(\textit{in days after last observation})}\bstrut\\\cline{2-4}
    Models &             1  &             7  &             14 \tstrut\bstrut\\
    \hline\hline
    Last                     &  0.851 (0.004) &  0.734 (0.002) &  0.691 (0.002) \tstrut \\
    Count                    &  0.860 (0.004) &  0.746 (0.002) &  0.699 (0.002) \\
    Ignore                   &  0.853 (0.003) &  0.740 (0.002) &  0.697 (0.002) \\
    Resample                 &  0.844 (0.004) &  0.739 (0.002) &  0.695 (0.002) \\
    \textbf{DeepJoint}         &  0.862 (0.003) &  0.742 (0.002) &  0.695 (0.002) \\
    GRU-D                    &  0.855 (0.004) &  0.743 (0.002) &  0.697 (0.002) \\
    Feature                  &  0.873 (0.003) &  0.750 (0.002) &  0.697 (0.002) \\
    \textbf{DeepJointFeature}  &  0.873 (0.003) &  0.749 (0.002) &  0.697 (0.002) \\
    \textbf{DeepJointFineTune} &  \textbf{0.878} (0.003) &  \textbf{0.751} (0.002) &  \textbf{0.700} (0.002) \\
    \end{tabular}
\end{table}

\begin{figure}[ht!]
\centering
\includegraphics[width=\textwidth]{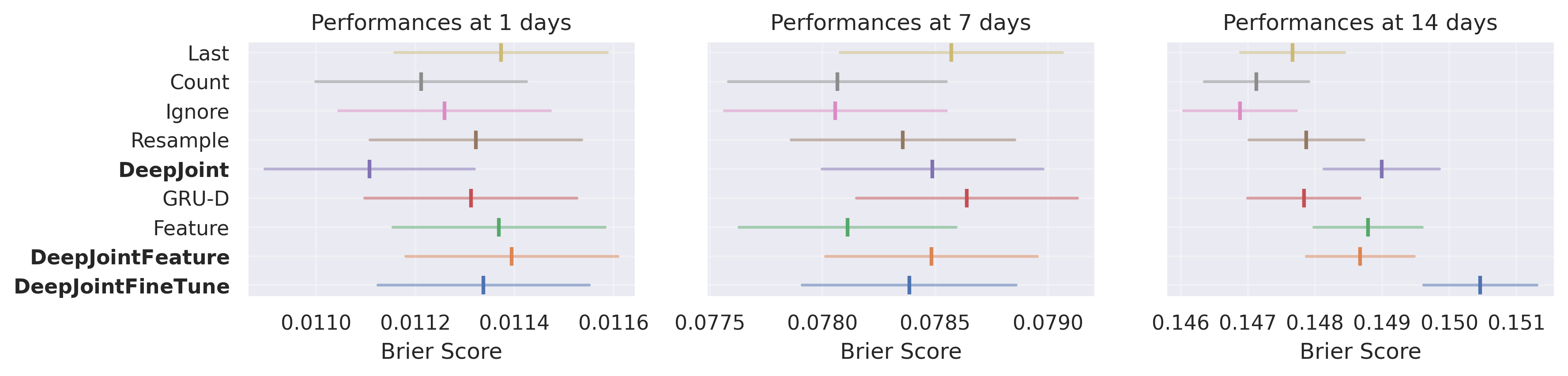}
\caption{Robustness evaluation between patients admitted on weekdays and weekends.}
\label{fig:brier}
\end{figure}

Figure \ref{fig:brier} and Table \ref{tab:brier} present the Brier score obtained on a random split of patients, showing little difference in the calibration of the different models.

\begin{table}[ht!]
    \centering
    \caption{Time dependent Brier score - Mean (95\%-CI) -- \textit{Lower is better.}}
    \label{tab:brier}
    \begin{tabular}{rccc}
    {} & \multicolumn{3}{c}{Evaluation horizons}\\
    {} & \multicolumn{3}{c}{(\textit{in days after last observation})}\bstrut\\\cline{2-4}
    Models &             1  &             7  &             14 \tstrut\bstrut\\
    \hline\hline
    Last                     &  0.011 (0.000) &  0.079 (0.000) &  0.148 (0.001) \tstrut\\
    Count                    &  0.011 (0.000) &  0.078 (0.000) &  0.147 (0.001) \\
    Ignore                   &  0.011 (0.000) &  0.078 (0.000) &  0.147 (0.001) \\
    Resample                 &  0.011 (0.000) &  0.078 (0.000) &  0.148 (0.001) \\
    \textbf{DeepJoint}         &  0.011 (0.000) &  0.078 (0.000) &  0.149 (0.001) \\
    GRU-D                    &  0.011 (0.000) &  0.079 (0.000) &  0.148 (0.001) \\
    Feature                  &  0.011 (0.000) &  0.078 (0.000) &  0.149 (0.001) \\
    \textbf{DeepJointFeature}  &  0.011 (0.000) &  0.078 (0.000) &  0.149 (0.001) \\
    \textbf{DeepJointFineTune} &  0.011 (0.000) &  0.078 (0.000) &  0.150 (0.001) \\
    \end{tabular}
\end{table}

\newpage
\subsubsection{Transfer performances}
\begin{figure}[ht!]
\centering
\includegraphics[width=\textwidth]{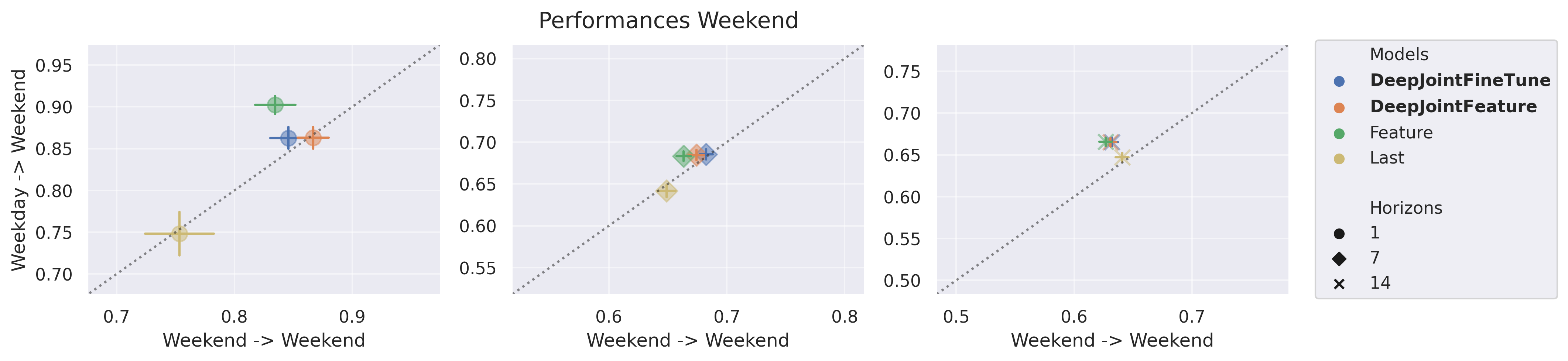}
\caption{Robustness evaluation between patients admitted on weekdays and weekends.}
\label{fig:split_weekend}
\end{figure}

Figure \ref{fig:split_weekend} shows the performance on patients admitted on weekends. This echoes the conclusion made in Section \ref{sec:rob} that the proposed model is more robust to change in the clinical presence.

Table \ref{tab:transfer} (resp. \ref{tab:transfer_we}) present the discriminative difference on the weekdays test set between the model trained on weekdays-admissions and the model transferred from weekends to weekdays (resp. on the weekends test set). In the context of weekdays performance, note that the proposed approach \textbf{DeepJoint} is more robust than the \textbf{Resample} approach but presents similar results than the \textbf{Ignore} model. Note that, for weekends evaluation, \textbf{GRU-D} presents state-of-the-art predictive performance and robustness at short time horizons but presents a larger confidence intervals. Additionally, this result didn't generalise to the other evaluation horizons. 

These results emphasise that the proposed joint modelling \textbf{DeepJointFeature} is more robust than state-of-the-art approaches. Interestingly, ignoring the clinical process (\textbf{Resample} or \textbf{Ignore}) seems to be less robust to a change in the observation process, echoing the remark made by \cite{lipton2016directly} about the difficulty of ignoring it. Finally, note how simple features such as in \textbf{Count} might be detrimental to the model due to a covariate shift -- particularly pronounced when applied to weekends admission.

\begin{table}[ht!]
    \centering
    \caption{Difference between model trained on weekends and model trained on weekdays tested on weekdays admission test set - Mean (95\%-CI) -- \textit{Lower absolute value is better.}}
    \label{tab:transfer}
    \begin{tabular}{rccc}
        {} & \multicolumn{3}{c}{Evaluation horizons}\\
        {} & \multicolumn{3}{c}{(\textit{in days after last observation})}\bstrut\\\cline{2-4}
        Models &             1  &             7  &             14 \tstrut\bstrut\\
        \hline\hline
            Last                     &   \textbf{0.001} (0.004) &  -0.010 (0.003) &  \textbf{-0.021} (0.002) \tstrut \\
            Count                    &  -0.061 (0.006) &  -0.036 (0.003) &  -0.030 (0.003) \\
            Ignore                   &  -0.016 (0.007) &  -0.021 (0.003) &  -0.035 (0.003) \\
            Resample                 &   0.042 (0.005) &  -0.006 (0.003) &  -0.031 (0.003) \\
            \textbf{DeepJoint}         &  -0.015 (0.006) &  -0.025 (0.003) &  -0.042 (0.003) \\
            GRU-D                    &  -0.015 (0.005) &  -0.029 (0.003) &  -0.033 (0.003) \\
            Feature                  &  -0.047 (0.006) &  -0.033 (0.003) &  -0.042 (0.003) \\
            \textbf{DeepJointFeature} &   0.007 (0.006) &   \textbf{0.004} (0.003) &  \textbf{-0.021} (0.003) \\
            \textbf{DeepJointFineTune} &  -0.019 (0.005) &  -0.009 (0.004) &  -0.035 (0.003) \\
    \end{tabular}
\end{table}

\begin{table}[ht!]
    \centering
    \caption{Difference between model trained on weekdays and model trained on weekends tested on weekends admission test set - Mean (95\%-CI) -- \textit{Lower absolute value is better.}}
    \label{tab:transfer_we}

    \begin{tabular}{rccc}
        {} & \multicolumn{3}{c}{Evaluation horizons}\\
        {} & \multicolumn{3}{c}{(\textit{in days after last observation})}\bstrut\\\cline{2-4}
        Models &             1  &             7  &             14 \tstrut\bstrut\\
        \hline\hline
        Last                     &  -0.005 (0.009) &  -0.007 (0.005) &  \textbf{0.007} (0.003) \tstrut\\
        Count                    &   0.130 (0.023) &  -0.001 (0.004) &  0.015 (0.004) \\
        Ignore                   &   0.060 (0.029) &  -0.006 (0.005) &  0.016 (0.003) \\
        Resample                 &   0.058 (0.012) &   0.005 (0.006) &  0.046 (0.004) \\
        \textbf{DeepJoint}         &   0.071 (0.011) &   0.014 (0.005) &  0.030 (0.003) \\
        GRU-D                    &   \textbf{0.003} (0.010) &   0.028 (0.004) &  0.039 (0.004) \\
        Feature                  &   0.068 (0.013) &   0.020 (0.004) &  0.039 (0.003) \\
        \textbf{DeepJointFeature}  &  -0.004 (0.002) &   0.010 (0.004) &  0.034 (0.004) \\
        \textbf{DeepJointFineTune} &   0.017 (0.003) &   \textbf{0.003} (0.004) &  0.033 (0.004) \\
    \end{tabular}
\end{table}

\end{document}